\pdfoutput=1

\documentclass[11pt]{article}

\usepackage[preprint]{acl}

\usepackage{times}
\usepackage{latexsym}

\usepackage[T1]{fontenc}

\usepackage[utf8]{inputenc}

\usepackage{microtype}

\usepackage{inconsolata}

\usepackage{graphicx}
\usepackage{amsmath}
\usepackage{enumitem}
\usepackage{algorithm}
\usepackage{algorithmic}
\usepackage{subcaption}
\usepackage{amssymb}
\usepackage{xcolor}

\title{TiSpell: A Semi-Masked Methodology for Tibetan Spelling Correction covering Multi-Level Error with Data Augmentation}
\author{
  \textbf{Yutong Liu\textsuperscript{1}},
  \textbf{Xiao Feng\textsuperscript{1}},
  \textbf{Ziyue Zhang\textsuperscript{1}},
  \textbf{Yongbin Yu\textsuperscript{1, *}},
  \textbf{Cheng Huang\textsuperscript{3}},
\\
  \textbf{Fan Gao\textsuperscript{1}},
  \textbf{Xiangxiang Wang\textsuperscript{1, *}},
  \textbf{Ban Ma-bao\textsuperscript{1}},
  \textbf{Manping Fan\textsuperscript{1}},
\\
  \textbf{Thupten Tsering \textsuperscript{1}},
  \textbf{Gadeng Luosang\textsuperscript{2}},
  \textbf{Renzeng Duojie\textsuperscript{2}},
  \textbf{Nyima Tashi\textsuperscript{2, *}},
\\
  \textsuperscript{1} School of Information and Software Engineering, University of Electronic Science and \\ Technology of China, \\
  \textsuperscript{2} School of Information Science and Technology, Tibet University, \\
  \textsuperscript{3} Department of Ophthalmology, University of Texas Southwestern Medical Center,
\\
  \small{
    \textbf{Correspondence:} \href{mailto:ybyu@uestc.edu.cn}{ybyu@uestc.edu.cn} \href{wxxlongtime@gmail.com}{wxxlongtime@gmail.com} \href{wxxlongtime@gmail.com}{niqiongda@163.com}
  }
}

\begin{document}
\maketitle
\begin{abstract}
Multi-level Tibetan spelling correction addresses errors at both the character and syllable levels within a unified model. Existing methods focus mainly on single-level correction and lack effective integration of both levels. Moreover, there are no open-source datasets or augmentation methods tailored for this task in Tibetan. To tackle this, we propose a data augmentation approach using unlabeled text to generate multi-level corruptions, and introduce TiSpell, a semi-masked model capable of correcting both character- and syllable-level errors. Although syllable-level correction is more challenging due to its reliance on global context, our semi-masked strategy simplifies this process. We synthesize nine types of corruptions on clean sentences to create a robust training set. Experiments on both simulated and real-world data demonstrate that TiSpell, trained on our dataset, outperforms baseline models and matches the performance of state-of-the-art approaches, confirming its effectiveness.
\end{abstract}

\section{Introduction}
The Tibetan language, with its unique orthographic and syntactic characteristics, faces significant challenges in the digital era. In Tibetan natural language processing (NLP), especially in the context of Large Language Models (LLMs) \cite{gao2025tlue, huang2025sunshine}, one of the major bottlenecks is the scarcity of large-scale, high-quality annotated corpora for spelling correction. Existing Tibetan datasets often rely heavily on manual verification, which is time-consuming, labor-intensive, and difficult to scale. As a result, there is a pressing need for automated spelling correction tools that can handle the complexities of Tibetan script efficiently and accurately.

In recent years, research efforts in Tibetan spelling detection and correction have been increasing. \citet{Liu2017, San2021} conducted statistical analyses of Tibetan syllable spelling errors. Rule-based approaches, such as the Tibetan Syllable Rule Model (TSRM) proposed by \citet{Zhu2014Check}, have been developed to identify syllable components and detect spelling errors. They also introduced a framework for automatic proofreading of Tibetan text and an algorithm to analyze connective relations \citep{Zhu2014Proofreading}. In statistical approaches, \citet{Tashi2022} combined Tibetan grammatical rules with statistical techniques to detect errors.

For deep learning-based methods, \citet{Hua2020} employed LSTM networks and demonstrated superior performance over traditional statistical models. \citet{Hua2021} utilized a Bi-LSTM in an encoder-decoder framework to correct errors in Tibetan verbs. \citet{Jiacuo2019BERT-BiLSTM} introduced an encoder-decoder model that combines BERT with Bi-LSTM to correct grammatical errors, achieving state-of-the-art (SOTA) results for Tibetan spelling correction. Despite these advances, the overall progress in this area remains limited, particularly with regard to deep learning approaches.

\begin{figure}[t]
  \centering
  \includegraphics[width=0.9\columnwidth]{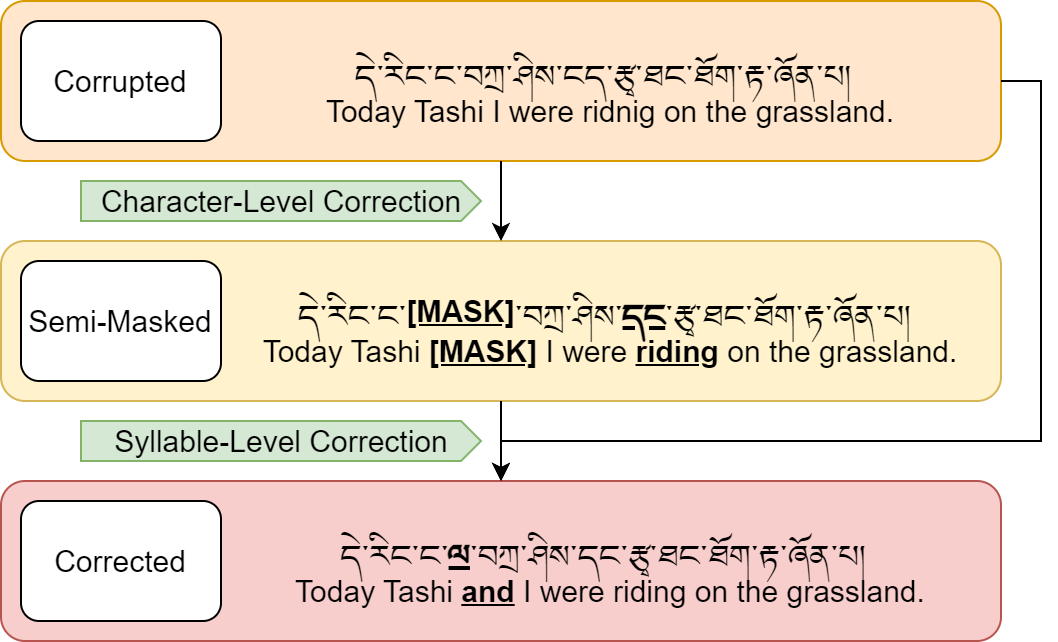} 
  \caption{The correction process of TiSpell.} 
  \label{fig:data_flow} 
\end{figure}

In contrast, spelling correction methods in other languages are more mature. Tools such as Hunspell \citep{hunspell}, SymSpell \citep{symspell}, and JamSpell \citep{jamspell}
are widely used. Deep learning models in other languages utilize diverse tokenization strategies. Early models focused on character-level tokenization for character-level error correction \citep{Kim2016CNNLSTM, Li2018LSTMLSTM}, while semi-character-level tokenization was introduced to mitigate local exchange errors \citep{Sakaguchi2016SCLSTM}. Subword tokenization methods—such as Unigram \citep{unigram}, BPE \citep{bpe}, WordPiece \citep{wordpiece}, and SentencePiece \citep{sentencepiece}—have become prevalent with the rise of general language models (GLMs) \citep{Sai2020Neuspell, Zhang2020SoftMaskedBERT, Martynov2024T5}.

The previously mentioned methods exhibit limited applicability, primarily addressing errors at the character or word level, and often lack interpretability. To address these challenges, we propose TiSpell, a novel Tibetan spelling correction model based on an encoder-only pre-trained language model (PLM) \citep{Liu2019RoBERTa, Yang2022CINO}. As shown in Figure~\ref{fig:data_flow}, the model comprises two correction heads: a character-level correction head, which corrects character-level errors and marks syllable-level errors with a mask token, and a syllable-level correction head, which recovers the masked tokens from the output of the character-level correction head. This design enables unified multi-level correction within a single PLM framework. Experimental results on both simulated and real-world data show that TiSpell outperforms previous models at both the character and syllable levels.
In summary, the key contributions of this work are as follows:
\begin{itemize}
    \item The data augmentation method utilizing non-labeled text is proposed to address multi-level Tibetan spelling corruptions and simulate diverse corruption intensity conditions.
    \item The Tibetan spelling correction benchmark is established, encompassing both traditional methods and deep learning models, which facilitates the evaluation and adaptation of methodologies originally developed for other languages.
    \item The multi-level Tibetan spelling correction model, TiSpell, has been developed. The implementation is available at https://github.com/Yutong-gannis/TiSpell.
\end{itemize}
\section{Related Works}
\subsection{Encoder-decoder Model}
Encoder-decoder model\citep{Cho2014EncoderDecoder, Uludogan2024EncoderDecoder} is the early architecture used for spelling correction. This type of architecture does not require a unified encode-decode tokenization. Correction models can encode using fine-grained tokenization and decode using word-level tokenization, allowing for more precise handling of detailed information\citep{Kim2016CNNLSTM, Li2018LSTMLSTM, Sakaguchi2016SCLSTM}. The encoder-decoder architecture can also facilitate the integration of RNN models with PLMs \citep{Sai2020Neuspell, Jiacuo2019BERT-BiLSTM}, such as BERT\citep{Devlin2019BERT} and ELMo\citep{Peters2018Elmo}. Transformer models also adopt the encoder-decoder architecture, where the encoder captures global information, while the decoder generates autoregressive outputs, making it particularly effective for long-term sequence generation\cite{Vaswani2023transformer, Martynov2024T5, Bijoy2025DPCSpell, Salhab2023AraSpell}. However, the encoder-decoder model consists of two separate components (encoder and decoder), increasing complexity and computational cost, which can also slow down inference time, especially in real-time applications.

\subsection{Encoder-Only Model}
Encoder-only model\cite{Devlin2019BERT, Liu2019RoBERTa, Peters2018Elmo, Yang2022CINO} originated from the encoder part of the transformer model\citep{Vaswani2023transformer}. This kind of architecture is more concise than the encoder-decoder model and still performs well in spelling correction. Soft-masked BERT \citep{Zhang2020SoftMaskedBERT} uses detection labels as soft masks on the input in BERT. 
Overfitting in error correction is one of the challenges that needs to be addressed\citep{Wu2023}. And encoder-only models use subword-level tokenization, which may result in the loss of character-level information.

\subsection{Detection-Correction Model}
Detection-correction model, a special case of the encoder-decoder model, is a more intuitively acceptable architecture that effectively decouples the process into two parts: detection and correction\citep{Li2024DeCoGLM, Bijoy2025DPCSpell, Zhang2020SoftMaskedBERT}. Previous research incorporates detection results as supplementary information to improve correction models\citep{Kaneko2020, Yuan2021}. They first perform detection and then fuse the detection results with text tokens in the correction model. \citet{Yuan2021} proposed the idea of using detection results as the mask of attention. DPCSpell\citep{Bijoy2025DPCSpell} and Soft-masked BERT\citep{Zhang2020SoftMaskedBERT} simply predict the error possibility in the detection model, while DeCoGML\citep{Li2024DeCoGLM} preserves the correct parts of the sentence when detecting errors.

\section{Augmentations Strategies}
Corrupting correct sentences is a common data augmentation strategy for spelling correction \citep{Sai2020Neuspell, Martynov2023Augmentation, Martynov2024T5}, enabling effective use of unlabeled text and reducing manual annotation. However, existing techniques, mainly designed for languages like English and Russian, often fail to account for the unique features of Tibetan. Tibetan is syllable-based, with homoglyph pairs and case-specific transformations that standard methods overlook. More details are introduced in Section~\ref{sec:appendix_tibetan}.

To address this, we propose a dual-level augmentation method tailored for Tibetan, introducing perturbations at both the character and syllable levels. We also adopt a mixed corruption strategy to simulate a broader range of real-world errors. The following section outlines our corruption techniques, with pseudo-code provided in Appendix~\ref{sec:appendix_augmentation}.

\subsection{Character-Level Corruption} 
Character-level corruption is designed to simulate errors caused by misreading, mishearing, and miscopying. NeuSpell\citep{Sai2020Neuspell} for English and Augmentex\cite{Martynov2023Augmentation} for Russian are notable examples of this kind of corruption. In this work, we define six types of character-level corruption:

\noindent\textbf{Random Deletion} refers to the process of randomly deleting characters from a sentence. Specifically, a syllable is first randomly selected, followed by the deletion of one character within that syllable.

\noindent\textbf{Random Insertion} refers to the process of randomly selecting a syllable in a sentence, choosing a character from the Tibetan alphabet, and inserting it into a random position within the selected syllable.

\noindent\textbf{Case Substitution} refers to the random selection of a character that is a root letter and has both upper and lower case forms and replacing it with its alternative case.

\noindent\textbf{Homoglyph Substitution} refers to randomly selecting a character that has more than one homoglyph and replacing it with one of its homoglyphs.

\noindent\textbf{Adjacent-Syllabic Character Transposition} refers to randomly select a syllable in sentence and randomly exchange positions of two characters within that syllable.

\noindent\textbf{Inter-Syllabic Character Transposition} refers to the process of randomly selecting two adjacent syllables in a sentence and swapping the positions of one character from each syllable.

\begin{figure}[t]
  \centering
  \includegraphics[width=0.6\columnwidth]{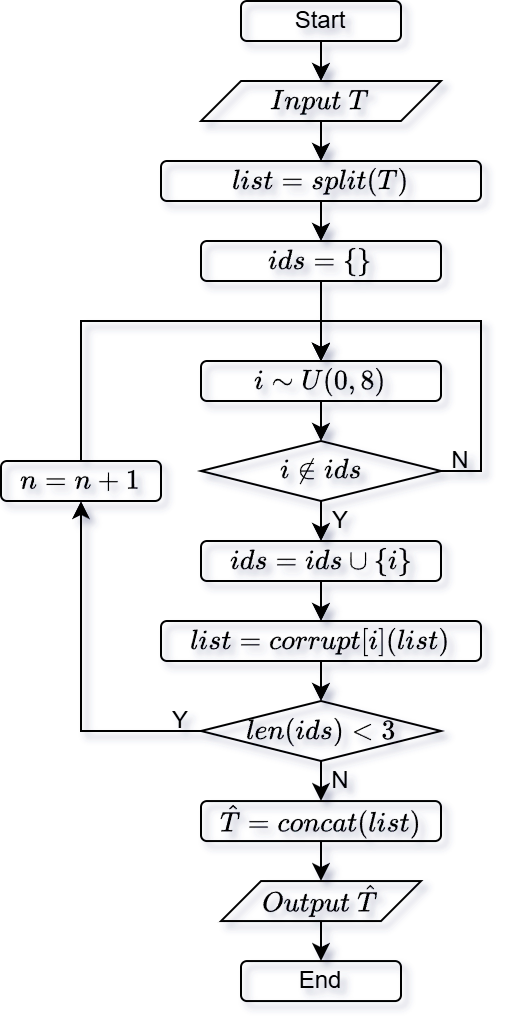}
  \caption{The flow chart of the mixed corruption strategy. The $corrupt$ is the set of the single corruption methods.}
  \label{fig:data_flow}
\end{figure}

\begin{figure*}[t]
  \centering
  \includegraphics[width=0.9\textwidth]{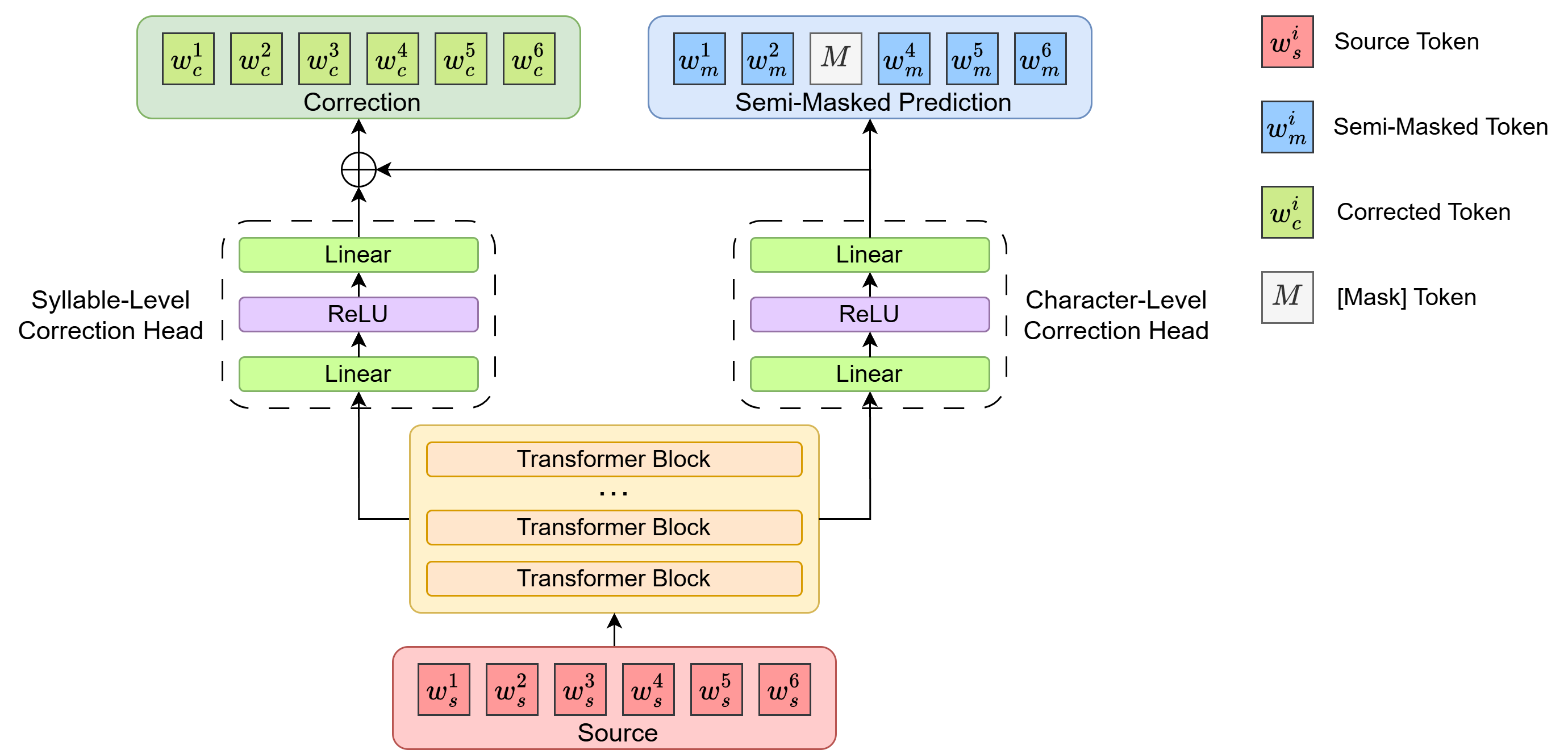}
  \caption{The proposed encoder-decoder structure based on PLM.}
  \label{fig:TiSpell}
\end{figure*}

\subsection{Syllable-Level Corruption}
The syllable-level errors are mainly caused by grammatical mistakes. Many grammatical error correction (GEC) methods \cite{Kaneko2020, Wu2023} use this kind of corruption. In this work, we define three types of syllable-level corruption:

\noindent\textbf{Random Deletion} refers to the random deletion of a syllable from a sentence. In our semi-masked approach, a semi-mask label is generated to preserve the correct portions of the text while replacing corrupted syllables with a [MASK] token. The purpose and application of this type of data will be discussed in Section~\ref{sec:character-level_correction_head}

\noindent\textbf{Random Transposition} refers to randomly select two syllables in a sentence and swap the positions of them.

\noindent\textbf{Random Merging} refers to randomly select two adjacent syllables in a sentence and merge them into a single syllable.

\subsection{Mixed Corruption}\label{sec:mix}

To simulate a harsher environment, this work proposes a mixed corruption strategy that combines multiple types of perturbations. In each instance, three out of the nine previously introduced corruption methods are randomly selected and applied. As illustrated in Fig.~\ref{fig:data_flow}, the Tibetan text $T$ is first split into a syllable list $\text{list}$, and an empty set $\text{ids} = \varnothing$ is initialized to keep track of the selected corruption method IDs. Then, a random integer $i \sim \mathcal{U}(0, 8)$ is drawn. If $i \notin \text{ids}$, the corresponding corruption method is applied to $\text{list}$, and $i$ is added to $\text{ids}$; otherwise, another $i$ is drawn. This process is repeated until three unique IDs have been collected in $\text{ids}$. Finally, the modified $\text{list}$ is concatenated to produce the corrupted text $\hat{T}$.

\section{Methods}
Our proposed model is an encoder-decoder model with two decoder heads, character-level and syllable-level correction head, learning from the advantage of the detection-correction model and leveraging the benefits of blank infilling pre-training\citep{Liu2019RoBERTa, Yang2022CINO}. The character-level correction head repairs the character-level errors and outputs a sentence with [MASK] tokens placed at the positions of potential error syllables, while the syllable-level correction head utilizes blank infilling to complete the semi-masked sentence. This section describes how we utilize the semi-mask output of character-level correction head and fusion of the character-level and syllable-level correction head, as depicted in Figure~\ref{fig:TiSpell}.

\subsection{Semi-Masked Character-Level Correction Head}\label{sec:character-level_correction_head}
The character-level correction process must accomplish two tasks: 1) correct character-level spelling errors, and 2) insert [MASK] tokens at positions where syllables are likely to have been deleted. 
Previous methods, such as Soft-masked BERT \cite{Zhang2020SoftMaskedBERT}, do not retain placeholders for deleted words, which can lead to confusion during prediction. In contrast, our method explicitly models deleted syllables by reserving their positions with [MASK] tokens at the character-level correction head.
The process of semi-masked character-level correction is shown as follows:

Given the sub-word tokens of correct text as:
\begin{equation}
  \label{eq:correct}
  w_\mathrm{c} = [w_\mathrm{c}^1, w_\mathrm{c}^2, \dots , w_\mathrm{c}^n]
\end{equation}
where $w_\mathrm{c}^i, i\in[1, n]$ is a single token of the text $w_\mathrm{c}$. The tokens of corrupted text can be represented as:

\begin{equation}
  \label{eq:source}
    w_\mathrm{s} = [w_\mathrm{s}^1, w_\mathrm{s}^2, \dots , w_\mathrm{s}^n]
\end{equation}

where $w_\mathrm{s}^j, j\in[1, n]$ is a single token of the text $w_\mathrm{s}$. The sub-word tokens of semi-masked text generated by augmentation methods can be shown as:

\begin{equation}
  \label{eq:detection_label}
  w_\mathrm{m}=\left[\begin{array}{c}
            w_\mathrm{m}^1 \\
            w_\mathrm{m}^2 \\
            \vdots \\
            w_\mathrm{m}^n
            \end{array}\right]^T,w_\mathrm{m}^k\in\{w_\mathrm{c}^k, M\}
\end{equation}

where $w_\mathrm{m}^k, k\in[1, n]$ is a token in semi-masked text $w_\mathrm{m}$, $M$ is the [mask] token. $w_\mathrm{m}^k=M$ means  a syllable selected in the $k$th position.

The proposed model begins by extracting the representations of the source text tokens by attention-based PLM as Equation~\ref{eq:plm}.
\begin{equation}
  \label{eq:plm}
  h = [h_1,h_2,\dots ,h_n] = PLM(w_\mathrm{s})
\end{equation}
where $h_i, i\in [1, n]$ is the hidden state of each token.
Within the PLM, the representation $h_i$ is computed using multi-head self-attention followed by a feed-forward network (FFN), defined as:

\begin{equation}
\label{eq:attention}
\text{Attention}(Q, K, V) = \text{softmax}\left(\frac{QK^\top}{\sqrt{d_k}}\right)V
\end{equation}

\begin{equation}
\label{eq:ffn}
\text{FFN}(x) = \text{ReLU}(xW' + b')W'' + b''
\end{equation}

where $Q$, $K$, and $V$ are the query, key, and value matrices projected from the input, and $W'$, $W''$, $b'$, $b''$ are learnable parameters.
The character-level corrected semi-masked predictions $\hat{w_\mathrm{d}}$ are generated through the character-level correction head, implemented by a fully connected network $f_\mathrm{C}$, as shown in Equation~\ref{eq:character_correction}: 
\begin{equation}
  \label{eq:character_correction}
  \hat{w}_\mathrm{m} = f_\mathrm{C}(h) = \text{ReLU}(hW_1 + b_1)W_2 + b_2
\end{equation}

where $\hat{w_\mathrm{d}}$ is the predicted output corresponding to $w_\mathrm{d}$. $W_1$, $W_2$, $b_1$, $b_2$ are learnable parameters in the character-level correction head $f_\mathrm{S}(h)$. In this step, character-level errors are corrected, and positions identified as potential syllable deletions are replaced with [MASK] tokens.

\subsection{Syllable-Level Correction Head}
The syllable-level correction process just needs to fill the [MASK] tokens with the correct tokens, similar to the pre-training approach of BERT\citep{Devlin2019BERT}. Comparing with existing methods, the task of syllable-level correction has been simplified as other errors have already been corrected in the character-level head.

The syllable-level correct predictions are generated through the syllable-level correction head, implemented by a fully-connected network $f_\mathrm{S}$.
The final result is to combine the prediction of character-level and syllable-level correction as presented in Equation~\ref{eq:complete}.
\begin{equation}
  \label{eq:complete}
  \hat{w_\mathrm{c}} = \hat{w_\mathrm{m}} + f_\mathrm{S}(h)
\end{equation}

\begin{equation}
  f_\mathrm{S}(h) = \text{ReLU}(hW_3 + b_3)W_4 + b_4
\end{equation}

where $\hat{w_\mathrm{c}}$ is the finally prediction of correction. $W_1$, $W_2$, $b_1$, $b_2$ are learnable parameters in the character-level correction head $f_\mathrm{S}(h)$.

\subsection{Multi-Task Training}\label{sec:multi_task_training}
The cross-entropy loss function, as illustrated in Equation~\ref{eq:loss}, is extensively employed in multi-label classification tasks and serves as the primary training objective for most error correction approaches.
\begin{equation}
  \label{eq:loss}
  L = -\frac{1}{n} \sum_{i=1}^n \sum_{j=1}^m y_{ij} \log(p_{ij})
\end{equation}
For multi-task learning, we utilize a weighted loss function to enable the model to concurrently acquire error detection and correction capabilities. The training objective for this TiSpell model is to minimize the loss function given by:
\begin{equation}
  \label{eq:multiloss}
  L = L_\mathrm{M} + w_\mathrm{C}\ L_\mathrm{C}
\end{equation}
where $L_\mathrm{M}$ is the average loss of multi-level correction prediction and $L_\mathrm{C}$ is the average loss of syllable-level semi-masked correction prediction. The syllable-level correction loss weight $w_\mathrm{C}$ is set to 2 to balance the scales of the two losses. For the impact of the loss weights on the model’s performance, please refer to Section~\ref{sec:analysis_weight}.

\section{Experiments}
\subsection{Dataset and Evaluation}\label{sec:dataset_evaluation}
\textbf{Dataset} The Tibetan University Sentiment Analysis (TUSA) dataset \citep{Zhu2023TUSA}, containing 10,000 sentences, is used for pre-training transformer-based models in Tibetan sentiment analysis. For spelling correction, the training data is sourced from the Tibetan text corpus released by Tibet University \citep{Zhang2022Dataset}, originally used for Tibetan news classification and comprising over 5.7 million sentences. From this corpus, 50,000 high-quality sentences are selected to construct the spelling correction training set. To assess model performance in real-world scenarios, an additional evaluation set of 1,000 erroneous sentences collected from the web is included. More detailed information can be found in Appendix~\ref{sec:appendix_dataset}.

\noindent\textbf{Setting} In the main experiment, we present the results of single systems trained on parallel data without any reranker. For models that require pre-training on a Tibetan corpus, the learning rate is initially set to $1 \times 10^{-4}$ during pre-training and is reduced to $5 \times 10^{-5}$ during fine-tuning. For RNN models, the learning rate is set to $5 \times 10^{-5}$ directly. The batch size is set to 128. Adam with Weight Decay (AdamW)\cite{Loshchilov2017AdamW} is used as the optimizer to improve model training stability and enhance generalization by incorporating weight decay during optimization. The value of weight decay is set to $1 \times 10^{-2}$ to avoid overfitting. More detailed information will be provided in Appendix~\ref{sec:appendix_setting}.

\begin{table*}[t!]
  \small
  \setlength{\tabcolsep}{4.5pt}
  \centering
  \begin{tabular}{l|ccc|ccc|ccc|ccc}
    \hline
     & \multicolumn{3}{c}{\textbf{Correct}} & \multicolumn{3}{|c}{\textbf{Char-Corrpt.}} & \multicolumn{3}{|c}{\textbf{Syll-Corrpt.}} & \multicolumn{3}{|c}{\textbf{Mix-Corrpt.}}\\
     
    \textbf{Model} & \textbf{P} & \textbf{R} & \textbf{F1} & \textbf{P} & \textbf{R} & \textbf{F1} & \textbf{P} & \textbf{R} & \textbf{F1} & \textbf{P} & \textbf{R} & \textbf{F1} \\
    \hline
    \multicolumn{13}{c}{\textbf{Traditional Method}} \\
    \hline
    dummy & \textbf{100.0} & \textbf{100.0} & \textbf{100.0} & 90.93 & 90.40 & 90.60 & - & - & - & 89.62 & 89.09 & 89.26 \\
    SymSpell & \textbf{100.0} & \textbf{100.0} & \textbf{100.0} & 90.93 & 90.43 & 90.60 & - & - & - & 89.62 & 89.09 & 89.26 \\
    HunSpell & \underline{99.55} & \underline{99.56} & \underline{99.55} & \underline{92.83} & \underline{92.47} & \underline{92.58} & - & - & - & \underline{91.70} & \underline{91.31} & \underline{91.43} \\
    JamSpell & 99.49 & 99.50 & 99.49 & \textbf{96.36} & \textbf{96.26} & \textbf{95.92} & - & - & - & \textbf{95.33} & \textbf{95.21} & \textbf{95.22} \\
    \hline
    \multicolumn{13}{c}{\textbf{Deep Learning Model}} \\
    \hline
    LSTM-LSTM & 96.53 & 96.78 & 96.47 & 92.85 & 93.45 & 92.83 & 65.93 & 69.09 & 66.77 & 76.51 & 78.68 & 76.96 \\
    CNN-LSTM & 97.17 & 97.33 & 97.07 & 93.34 & 93.89 & 93.10 & 65.25 & 68.12 & 65.99 & 76.15 & 78.19 & 76.56 \\
    SC-LSTM & 95.37 & 95.67 & 95.33 & 90.40 & 91.02 & 90.34 & 64.54 & 67.67 & 65.32 & 73.60 & 75.84 & 74.07 \\
    BERT-char & 96.26 & 97.04 & 96.52 & 90.82 & 91.97 & 91.14 & 60.69 & 63.16 & 60.69 & 70.82 & 73.42 & 71.55 \\
    BERT & 98.87 & 98.88 & 98.86 & 96.67 & 97.71 & 97.66 & 90.93 & 91.39 & 91.08 & 92.23 & 92.60 & 92.34 \\
    CINO & 98.55 & 98.59 & 98.55 & 97.14 & 97.22 & 97.15 & 89.31 & 89.90 & 89.50 & 90.37 & 90.94 & 90.56 \\
    RoBERTa & 99.07 & 99.08 & 99.06 & 97.86 & 97.89 & 97.85 & 90.44 & 90.96 & 90.63 & 92.15 & 92.56 & 91.28 \\
    RoBERTa + Bi-LSTM & 98.90 & 98.88 & 98.88 & 97.78 & 97.83 & 97.78 & \textbf{91.93} & \underline{92.39} & \underline{92.09} & 92.41 & 92.87 & 92.57 \\
    Soft-Masked RoBERTa & 98.91 & 98.96 & 98.92 & \underline{98.00} & \underline{98.06} & \underline{98.00} & 91.52 & 92.36 & 91.85 & \underline{92.42} & \underline{93.01} & \underline{92.64} \\
    TiSpell-CINO* & \underline{99.09} & \underline{99.11} & \underline{99.09} & 97.92 & 97.96 & 97.92 & 90.36 & 91.20 & 90.68 & 92.22 & 92.79 & 92.43 \\
    TiSpell-RoBERTa* & \textbf{99.25} & \textbf{99.26} & \textbf{99.22} & \textbf{98.30} & \textbf{98.36} & \textbf{98.31} & \underline{91.82} & \textbf{92.62} & \textbf{92.13} & \textbf{93.75} & \textbf{93.24} & \textbf{93.43} \\
    \hline
  \end{tabular}
  \caption{\label{tab:main_result}
    Percentage results on Tibetan multi-level spelling correction benchmarks. The parameter (\textbf{Param.}) counts of the backbones of each model are shown in the second column. The table shows the average precision (\textbf{P}), recall (\textbf{R}), and f1 score (\textbf{F1}) of models working on correct text (\textbf{Correct}), character-level corruption (\textbf{Char-Corrpt.}), syllable-level corruption (\textbf{Syll-Corrpt.}), and mixed corruption (\textbf{Mix-Corrpt.}). The highest metric is indicated in bold, while the second highest metric value is underlined. The "*" symbol denotes the models proposed in this study and the "-" symbol reflects the method's limitation in correcting corruption at the current level.}
\end{table*}

\noindent\textbf{Evaluation} 11 kinds of source text are used in the evaluation. Correct text, serving as the control group, is employed to assess the model's ability to refrain from making changes when the input sentence is error-free. Six types of character-level and three types of syllable-level corrupted text are utilized to evaluate the model's ability to correct a single spelling error. Mixed corrupted text, described in Section~\ref{sec:mix}, is introduced to test the model's performance under more rigorous and complex conditions. Three types of metrics are used for evaluation, precision, recall, and F1 score\citep{Powers2011Metrics}. For an intuitive comparison of character-level and syllable-level correction performance, we report the mean precision, recall, and F1 scores of correction on each level in the main results.

\subsection{Model Settings}
\textbf{Proposed Models} Two open-source PLM models, CINO-base-v2\citep{Yang2022CINO} and RoBERTa-base \citep{Liu2019RoBERTa}, both have been pre-trained on Tibetan corpora, are utilized as the backbones for TiSpell. The semi-masked character-level correction head and the syllable-level correction head each employ a fully connected network architecture with two hidden layers, where dimension of the hidden layers aligns with the hidden state size of the PLM. 

\noindent\textbf{Baseline Models} In traditional methods, the dummy is used to simulate the scenario where no correction is applied. SymSpell\cite{symspell}, HunSpell\cite{hunspell} and JamSpell\cite{jamspell}  are chosen as representative methods. 
A Tibetan corpus consisting of 1,000,000 sentences, selected from the training dataset mentioned above \citep{Zhang2022Dataset}, is used to evaluate the performance of JamSpell. 

In deep learning models, we select CNN-LSTM\cite{Kim2016CNNLSTM}, LSTM-LSTM\cite{Li2018LSTMLSTM}, SC-LSTM\cite{Sakaguchi2016SCLSTM}, RoBERTa+Bi-LSTM\cite{Jiacuo2019BERT-BiLSTM} as encoder-decoder architectures. The first three models incorporate four LSTM layers, while the RoBERTa+Bi-LSTM architecture employs two bidirectional LSTM layers, consistent with the design specified in its foundational work. In particular, RoBERTa + Bi-LSTM \cite{Jiacuo2019BERT-BiLSTM} has demonstrated superior performance, establishing itself as the SOTA model for Tibetan spelling correction tasks prior to recent advancements.
BERT\cite{Devlin2019BERT}, RoBERTa\cite{Liu2019RoBERTa}, and CINO\cite{Yang2022CINO} are three encoder-only models for comparison. To investigate the impact of character-level tokenization on encoder-only architectures, we implement a BERT model specifically designed for character-level tokenization. Soft-Masked RoBERTa\cite{Zhang2020SoftMaskedBERT} represents a detection-correction framework that has demonstrated superior performance in Chinese spelling correction tasks, as evidenced by previous research. More detailed information will be provided in Appendix~\ref{sec:appendix_setting}.

\subsection{Main Results}
\textbf{Augmented Dataset}
Table~\ref{tab:main_result} presents the performance of various Tibetan spelling correction methods. Traditional rule-based approaches perform adequately on clean text but show clear limitations: HunSpell and SymSpell are restricted to character-level corrections, while JamSpell, despite using statistical context, struggles with syllable-level errors.

Deep learning models using character- or semi-character-level tokenization also underperform on syllable-level errors, highlighting the limitations of these tokenization schemes. For example, BERT-char performs 1.5 F1 points worse than standard BERT, indicating a suboptimal balance between complexity and effectiveness.

In contrast, our proposed multi-level correction models based on CINO and RoBERTa achieve consistent improvements, with an average 1.5 F1 gain in syllable-level correction without increasing model parameters. TiSpell-RoBERTa, in particular, ranks among the top two models in four evaluation settings, matching or surpassing both Soft-Masked RoBERTa and the SOTA RoBERTa + Bi-LSTM model. Detailed results by corruption type are available in Appendix~\ref{sec:appendix_metrics}.

\begin{table}[h]
  \small
  \centering
  \begin{tabular}{c|ccc}
    \hline
     \textbf{Model} & \textbf{P} & \textbf{R} & \textbf{F1}\\
    \hline
    LSTM-LSTM & 74.83 & 75.45 & 74.67 \\
    CNN-LSTM  & 74.29 & 74.91 & 74.56 \\
    SC-LSTM   & 72.78 & 74.32 & 73.64 \\
    BERT      & 90.88 & 91.13 & 91.70 \\
    CINO      & 88.25 & 88.96 & 87.41 \\
    RoBERTa   & 91.59 & 91.36 & 90.87 \\
    RoBERTa + Bi-LSTM   & 91.19 & 91.52 & \underline{91.61} \\
    Soft-Masked RoBERTa & \underline{91.48} & \textbf{92.73} & 91.27 \\
    TiSpell-CINO*    & 91.39 & 91.84 & 91.11 \\
    TiSpell-RoBERTa* & \textbf{92.66} & \underline{91.93} & \textbf{92.34} \\
    \hline
  \end{tabular}
  \caption{\label{tab:realworld}
    Evaluation on real-world dataset.}
\end{table}

\noindent\textbf{Real-World Dataset}
Table~\ref{tab:realworld} shows the evaluation on a real-world Tibetan dataset. TiSpell-RoBERTa achieves the best overall performance with an F1 score of 92.34, outperforming strong baselines like RoBERTa + Bi-LSTM and Soft-Masked RoBERTa. It also obtains the highest precision (92.66), indicating fewer false corrections. While Soft-Masked RoBERTa has slightly higher recall, its overall F1 is lower. These results highlight the robustness of our multi-level model in real-world scenarios, effectively handling both character- and syllable-level errors.

\section{Analysis}

\subsection{Weights of Multi-Task Loss}\label{sec:analysis_weight}
In Section~\ref{sec:multi_task_training}, we introduced a multi-task training strategy to balance the two prediction tasks. To determine the character-level loss weight $w_C$ which significantly influences the training objective in Equation~\ref{eq:multiloss}, we conducted preliminary experiments using the TiSpell-RoBERTa model. Based on an initial analysis of the loss scale, we set $w_C = 0.5$ as the baseline and evaluated a range of values ${0.5, 0.8, 1, 2, 5}$ to investigate their effects. The results are presented in Table~\ref{tab:weight_result}. Our experiments demonstrate that the overall correction loss plays a slightly more significant role than the semi-masked correction loss, with $w_C = 2$ achieving optimal performance for our task.

\begin{table}[h]
  \small
  \centering
  \begin{tabular}{c|ccc}
    \hline
     & \multicolumn{3}{c}{\textbf{F1 score}} \\
     \textbf{$w_C$} & \textbf{Char-Corrpt.} & \textbf{Syll-Corrpt.} & \textbf{Mix-Corrpt.}\\
    \hline
    0.5 & \underline{98.16} & \underline{92.00} & \underline{93.20} \\
    0.8 & 97.86 & 91.68 & 92.57 \\
    1.0 & 98.04 & 91.67 & 92.89 \\
    2.0 & \textbf{98.31} & \textbf{92.13} & \textbf{93.43} \\
    5.0 & 97.64 & 90.89 & 92.12 \\
    \hline
  \end{tabular}
  \caption{\label{tab:weight_result}
    Impact of loss weight variations on model performance: F1 score (\%) evaluation. $w_C$ is defined in Section~\ref{sec:multi_task_training}}
\end{table}

\begin{table*}[t!]
  \small
  \setlength{\tabcolsep}{2pt}
  \centering
  \begin{tabular}{c|cccc|cccc}
    \hline
     \textbf{Backbone} & \textbf{Pretrained} & \textbf{Multi-Head} & \textbf{Res. Connect} & \textbf{Head Layer} & \textbf{Correct} & \textbf{Char-Corrpt.} & \textbf{Syll-Corrpt.} & \textbf{Mix-Corrpt.}\\
    \hline
    RoBERTa & \checkmark & \checkmark & \checkmark & 2 & \textbf{99.22} & \textbf{98.31} & \textbf{92.13} & \textbf{93.43} \\
    RoBERTa & \checkmark & \checkmark & \checkmark & 1 & 98.90 & 97.99 & 91.44 & 92.87 \\
    RoBERTa & \checkmark & \checkmark & - & 2 & 98.89 & 97.85 & 91.41 & 92.23 \\
    RoBERTa & \checkmark & - & - & 2 & 99.06 & 97.85 & 90.63 & 91.28 \\
    RoBERTa & - & \checkmark & \checkmark & 2 & 98.82 & 97.49 & 90.05 & 91.51 \\
    \hline
    CINO & \checkmark & \checkmark & \checkmark & 2 & \textbf{99.09} & \textbf{97.92} & \textbf{90.68} & \textbf{92.43} \\
    CINO & \checkmark & \checkmark & \checkmark & 1 & 99.03 & 97.74 & 89.92 & 91.28 \\
    CINO & \checkmark & \checkmark & - & 2 & 98.87 & 97.73 & 90.29 & 91.41 \\
    CINO & \checkmark & - & - & 0 & 98.55 & 97.15 & 89.50 & 90.56 \\
    \hline
  \end{tabular}
  \caption{\label{tab:ablation_result}
    Ablation study results. The table shows the average F1 score (\%) of models working on correct text, character-level corruption, syllable-level corruption, and mixed corruption.}
\end{table*}

\subsection{Ablation Study}
To assess the contributions of key components in our model, we perform an ablation study on the multi-head architecture, head layers, and residual connection, as shown in Table~\ref{tab:ablation_result}.

\noindent\textbf{Semi-masked character-level correction head}
The semi-masked character-level correction head is a key feature of our model. By removing this head (row 4), we observe a significant drop in syllable-level correction performance, which highlights its importance in simplifying syllable-level tasks and improving overall correction effectiveness. This result underscores the critical role of the semi-masked head in achieving accurate and robust text correction.

\noindent\textbf{Effectiveness of head layers}
Each head in our model shares features extracted from the PLM network, with the number of layers influencing the model's ability to refine and process these features. Comparing the performance between the first and second rows of Table~\ref{tab:ablation_result}, we find that increasing the number of layers to two per head results in a stable and consistent improvement in performance. This demonstrates that deeper head layers help the model better utilize shared features, leading to superior correction outcomes.

\noindent\textbf{Residual connection}
The residual connection between heads plays a crucial role in refining the syllable-level correction process. By enabling the syllable-level head to focus on recovering deleted syllables more effectively, the residual connection enhances model performance. A comparison of the first and third rows of Table~\ref{tab:ablation_result} shows that the introduction of the residual connection leads to a measurable improvement in correction accuracy, emphasizing its significance in optimizing the model’s performance.

\subsection{Interpretability Analysis}
\begin{figure}[h]
  \centering
  \includegraphics[width=0.9\columnwidth]{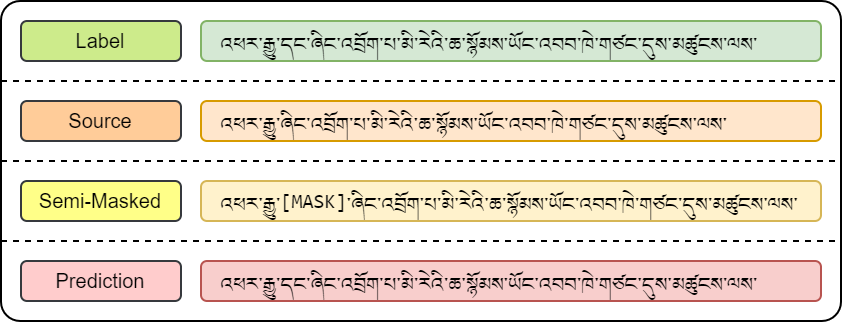}
  \caption{A Prediction Example of Tispell-RoBERTa.}
  \label{fig:example}
\end{figure}

To validate the reliability of the proposed methodology, we perform a comprehensive interpretability analysis with a focus on the attention mechanisms in PLM. 
Attention mechanisms play a crucial role in capturing contextual dependencies and improving model performance. To gain deeper insights into the model's information-extraction process, we analyze a sample case involving the deletion of a syllable in the input text, as illustrated in Figure~\ref{fig:example}, and visualize the dot-product attention scores. These scores reveal the relative importance assigned to different input tokens during prediction.

\begin{figure}[h!]
    \centering
    \begin{subfigure}[b]{\columnwidth}
        \centering  
        \includegraphics[width=0.85\textwidth]{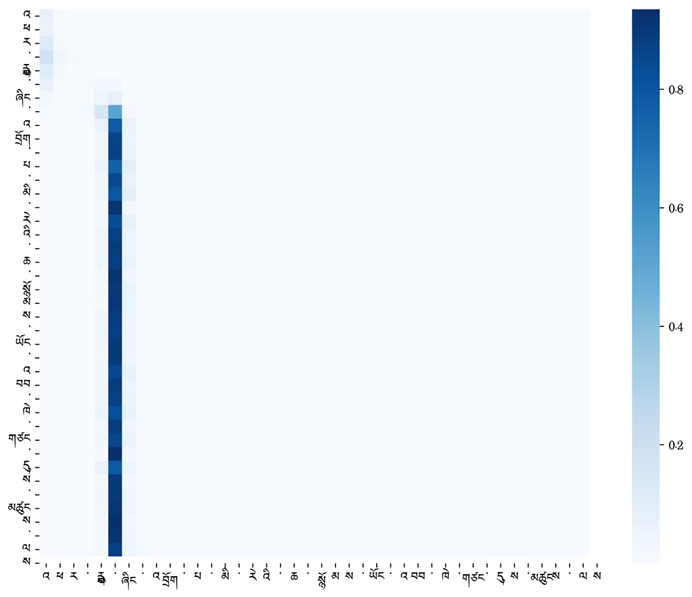}  
        \caption{2$th$ Layer}  
        \label{early_layer}  
    \end{subfigure}  
    \hfill
    \begin{subfigure}[b]{\columnwidth}  
        \centering
        \includegraphics[width=0.85\textwidth]{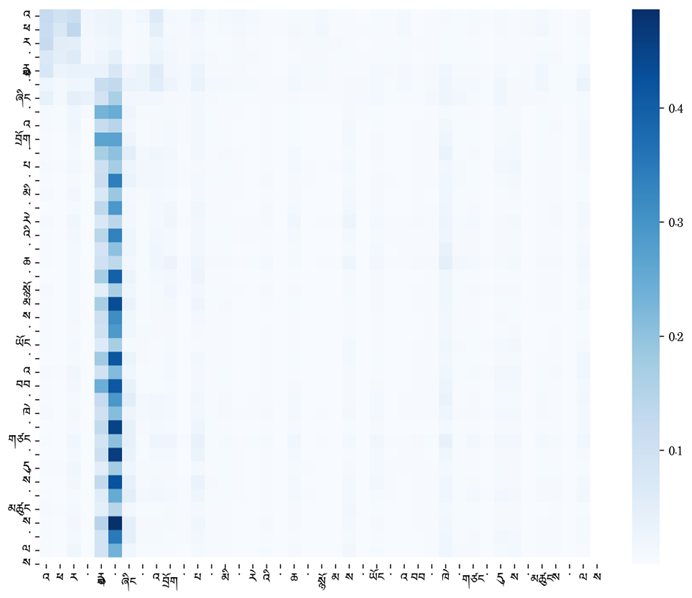}  
        \caption{8$th$ Layer}  
        \label{last_layer}
    \end{subfigure}
    \caption{Visualization of Attention Score in the early and late layer in the proposed model}
    \label{fig:attention_heatmap}  
\end{figure}

Figure~\ref{fig:attention_heatmap} visualizes the attention scores from the first head of both the second and eighth layers, providing insights into the model's focus at different stages of processing. As shown, the model assigns higher attention scores to the position where a syllable is missing, while paying less attention to the surrounding correct tokens. This behavior demonstrates the model's capability to identify and focus on locations where syllables need to be added.

The visualization further reveals that the model consistently attends to tokens containing errors, while other tokens tend to exhibit identity mapping. This pattern aligns with the characteristic of the spelling error correction task, where the model prioritizes correcting erroneous tokens while preserving the correctness of unaffected tokens.

\section{Conclusion}
We propose a semi-masked methodology for multi-level Tibetan spelling correction. This framework incorporates a semi-masked character-level correction head as an auxiliary task to streamline syllable-level correction and enhance overall correction performance. A multi-task learning strategy is employed to integrate both functionalities within a unified model, optimized using a weighted loss function.
Experimental results on both simulated and real-world data demonstrate that our proposed model, TiSpell, outperforms previous encoder-decoder architectures and achieves F1 scores comparable to state-of-the-art (SOTA) models on Tibetan spelling correction benchmarks. Ablation studies further validate the effectiveness of our model design. We hope this work will provide valuable insights and guide future research in Tibetan spelling correction, particularly in addressing multi-level errors.

\section{Limitations}
Despite the comprehensive evaluation conducted in this study, there are several limitations that should be acknowledged. First, due to resource and time constraints, we were unable to reproduce the results of some existing models in the field (e.g., DPCSpell\citep{Bijoy2025DPCSpell}, DeCoGLM\cite{Li2024DeCoGLM}), which limits the direct comparability of our approach with the full spectrum of SOTA methods. Second, limited by GPUs computing ability, our work does not explore spelling correction techniques based on large language models (e.g., Llama, T5\cite{Martynov2024T5}), which have shown promising results in various NLP tasks.

\section*{Ethics Statement}
In conducting our research on Tibetan spelling correction, we acknowledge the critical importance of addressing potential ethical implications and ensuring the responsible deployment of the developed technology. To uphold ethical standards throughout the study, we have implemented the following measures:

\noindent\textbf{Dateset and Model.}
The datasets and models we used are publicly available and utilized only for research purposes. The datasets do not contain any information that names or uniquely identifies individual people or offensive content.

\noindent\textbf{Nationality Equality.}
Our research is dedicated to Tibetan, one of the minority languages in China, and is firmly grounded in the principle of ethnic equality. The datasets used in this study have been carefully curated to eliminate any content that could promote discrimination against ethnic minorities or undermine national unity, ensuring that our work aligns with ethical and inclusive research practices.

\bibliography{custom}

\appendix
\section{Structure of Tibetan Text}\label{sec:appendix_tibetan}
A syllable is the fundamental unit of meaning in Tibetan. Influenced by cultural and historical factors, Tibetan syllables can be broadly categorized into two types: native Tibetan syllables and transliterated syllables. Native syllables adhere to the conventional rules of Tibetan syllable formation, while transliterated syllables are derived from Sanskrit, primarily found in Buddhist scriptures.

Tibetan syllables are constructed in a two-dimensional arrangement using Tibetan characters, which include 30 consonants, 4 vowels, and 20 numerals. The structure of a native Tibetan syllable typically comprises seven types of letters: prefix, root, superscript, subscript, vowel, suffix, and farther suffix. Among these, the upright unit—consisting of the root letter, superscript, subscript, and vowel—forms the vertical core of the syllable.

\section{Dataset Details}\label{sec:appendix_dataset}

\begin{table}[h]
  \small
  \setlength{\tabcolsep}{1.5pt}
  \centering
  \begin{tabular}{l|c|c|c|c|c}
    \hline
     \textbf{class} & \textbf{Size} & \textbf{Avg. Char.} & \textbf{Avg. Syll.} & \textbf{75\% Syll.} & \textbf{95\% Syll.} \\
    \hline
    \multicolumn{5}{c}{Tibetan news classification dataset} \\
    \hline
    Eco. & 351723 & 95.0 & 23.9 & 32.0 & 62.0\\
    Cult. & 210003 & 70.4 & 18.4 & 21.0 & 46.0 \\
    Edu. & 426594 & 89.5 & 22.6 & 29.0 & 59.0 \\
    Sci. & 207070 & 85.1 & 21.6 & 28.0 & 55.0 \\
    Jour. & 152724 & 85.5 & 21.8 & 26.0 & 53.0 \\
    Art. & 477524 & 66.7 & 17.3 & 22.0 & 43.0 \\
    Gov. & 1365724 & 106.7 & 26.5 & 35.0 & 71.0 \\
    Life. & 613417 & 71.3 & 18.3 & 23.0 & 48.0 \\
    Pol. & 911602 & 98.0 & 22.6 & 33.0 & 66.0 \\
    Law. & 995968 & 90.0 & 22.6 & 29.0 & 58.0\\
    \hline
    \multicolumn{5}{c}{TUSA} \\
    \hline
    Pos. & 4836 & 120.2 & 34.0 & 41.0 & 73.0 \\
    Neg. & 4856 & 102.2 & 29.0 & 35.0 & 90.25\\
    \hline
  \end{tabular}
  \caption{\label{tab:dataset_detail} The description analysis of the Tibetan news classification dataset and TUSA dataset. Abbreviations used in the table are as follows: Eco (Economy), Cult (Culture), Edu (Education), Sci (Science), Jour (Journey), Art (Art), Gov (Government), Life (Livelihood), Pol (Politics), Law (Law), Pos (Positive), Neg (Negative).}
\end{table}

The information of the datasets for training is presented in Tables~\ref{tab:dataset_detail}. The Tibetan news classification dataset comprises ten categories, including Economy, Culture, Education, Science, and others, with sample sizes ranging from approximately 150,000 to over 1.3 million. Each category exhibits different linguistic characteristics for instance, 'Government' and 'Politics' tend to have longer texts, averaging over 25 syllables per sentence, while categories like 'Art' and 'Culture' contain shorter entries.

In addition, the Tibetan sentiment analysis dataset (TUSA) includes two classes: Positive and Negative, each with around 4,800 samples. Sentiment samples are generally longer, with average syllable counts exceeding those in most news categories. The '75\%' and '95\%' quantile syllable lengths further illustrate the distribution spread, indicating a considerable variation in text length that poses challenges for sequence modeling. Overall, the datasets represent a diverse and realistic collection of Tibetan text, supporting robust model training and evaluation.

\section{Setting Details}\label{sec:appendix_setting}

\begin{table}[h]
  \small
  \setlength{\tabcolsep}{2pt}
  \centering
  \begin{tabular}{l|cc}
    \hline
    \textbf{Configuration} & \textbf{RoBERTa} & \textbf{CINO} \\ \hline
    Model Type & RoBERTa & XLM-RoBERTa \\ \hline
    Attention Dropout & 0.1 & 0.1 \\ \hline
    BOS Token ID & 0 & 0 \\ \hline
    EOS Token ID & 2 & 2 \\ \hline
    Hidden Activation & GELU & GELU \\ \hline
    Hidden Dropout & 0.1 & 0.1 \\ \hline
    Hidden Size & 768 & 768 \\ \hline
    Initializer Range & 0.02 & 0.02 \\ \hline
    Intermediate Size & 3072 & 3072 \\ \hline
    Layer Norm Epsilon & 1e-12 & 1e-05 \\ \hline
    Max Position Embeddings & 512 & 514 \\ \hline
    Number of Attention Heads & 12 & 12 \\ \hline
    Number of Hidden Layers & 12 & 12 \\ \hline
    Pad Token ID & 1 & 1 \\ \hline
    Position Embedding Type & Absolute & Absolute \\ \hline
    Type Vocab Size & 2 & 1 \\ \hline
    Use Cache & True & True \\ \hline
    Vocabulary Size & 8094 & 8094 \\ \hline
  \end{tabular}
  \caption{\label{tab:model_config} Model Configurations for RoBERTa and CINO}
\end{table}

We utilize the entire set of sentences from the TUSA dataset for pre-training, and select 500,000 sentences from the Tibetan news classification dataset for fine-tuning. The news classification dataset is partitioned using a split ratio of 99:1. The training process was carried out on an RTX 4090 GPU with 24GB of memory. The average time consumed per epoch for the TiSpell-RoBERTa model is approximately 52 minutes. All models underwent a two-phase training process, consisting of 30 epochs of pre-training followed by 60 epochs of fine-tuning.

According to the average number of characters and the 95th percentile of syllables presented in Table~\ref{tab:dataset_detail}, the output length for character-level tokenization models is configured at 256, while for syllable-level tokenization models, it is set to 96. The hidden size of the LSTM models is set to 512. And the configurations of RoBERTa and CINO are shown in Table~\ref{tab:model_config}.

\section{Metrics Details}\label{sec:appendix_metrics}
\begin{figure}[h]
  \includegraphics[width=\columnwidth]{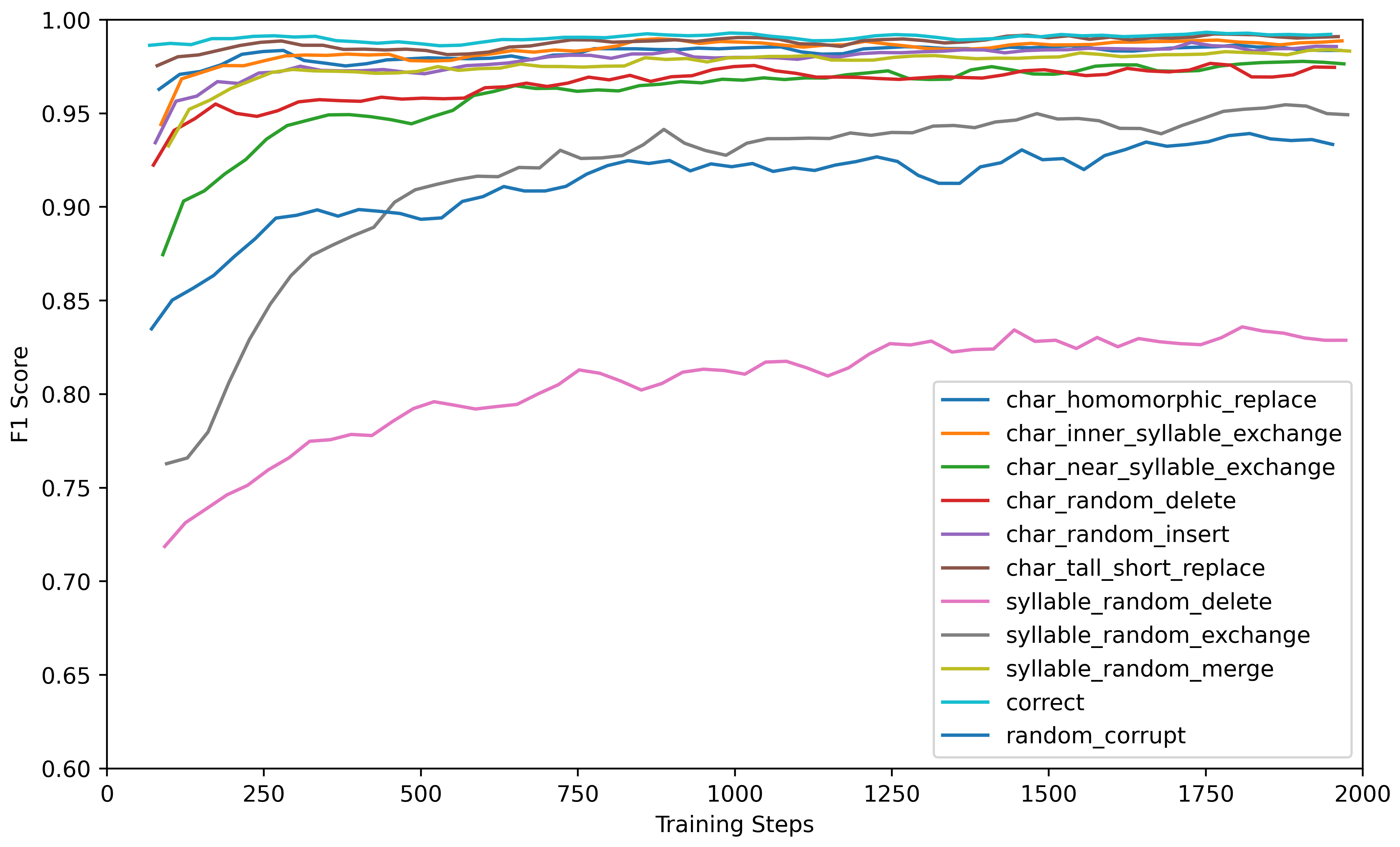}
  \caption{The metrics curves of TiSpell-RoBERTa.}
  \label{fig:step_result}
\end{figure}

The F1 score curves for all types of corruptions in TiSpell-RoBERTa during training steps are shown in Figure~\ref{fig:step_result}. As depicted, character-level correction consistently outperforms syllable-level correction across all metrics. Additionally, all metrics have converged during training.

The final F1 scores of all types of corruptions for TiSpell-RoBERTa and TiSpell-CINO are shown in Table~\ref{tab:tispell_metrics}. TiSpell-RoBERTa outperforms TiSpell-CINO in most corruption types, especially in character-level and syllable-level errors, with notable improvements in random deletion and transposition.

\begin{table}[h]
  \small
  \setlength{\tabcolsep}{3pt}
  \centering
  \begin{tabular}{l|cc}
    \hline
    \textbf{Metrics} & \textbf{TiSpell-RoBERTa} & \textbf{TiSpell-CINO} \\ 
    \hline
    Correct & 99.22 & 99.09 \\
    Rand. Corrupt & 93.43 & 92.43 \\
    \hline
    \multicolumn{3}{c}{Character Level} \\
    \hline
    Rand. Del. & 97.31 & 96.53 \\
    Rand. Ins. & 98.52 & 98.21 \\
    Case. Sub. & 99.12 & 98.87 \\
    Homo. Sub. & 98.46 & 98.09 \\
    Adj.-Syll. Trans. & 97.63 & 97.26 \\
    Inter-Syll. Trans. & 98.81 & 98.52 \\
    \hline
    \multicolumn{3}{c}{Syllable Level} \\
    \hline
    Rand. Del. & 83.07 & 81.04 \\
    Rand. Trans. & 95.08 & 92.87 \\
    Rand. Merge. & 98.25 & 98.13 \\
  \end{tabular}
  \caption{\label{tab:tispell_metrics} F1 Scores Across All Corruption Types for TiSpell-RoBERTa and TiSpell-CINO. Abbreviations: Rand. (Random), Char. (Character), Del. (Deletion), Ins. (Insertion), Sub. (Substitution), Homo. (Homoglyph), Syll. (Syllabic), Trans. (Transposition), Adj. (Adjacent).}
\end{table}

\section{Augmentation details}\label{sec:appendix_augmentation}
Character-level random deletion involves randomly removing characters from a sentence. The process begins by selecting a syllable at random, then deleting a single character within that syllable. The pseudo-code is presented in Algorithm~\ref{alg:char_random_delete}.
\begin{algorithm}
    \caption{char\_random\_delete}\label{alg:char_random_delete}
    \begin{algorithmic}[1]
    \STATE \textbf{Input:} List of syllables \texttt{syllables}
    \STATE Randomly select an index: \texttt{i} = Random(0, len(syllables)-1)
    \STATE Delete element at \texttt{i}: \texttt{syllables.delete(i)}
    \STATE \textbf{Return} \texttt{syllables}
    \end{algorithmic}
\end{algorithm}

Character-level random insertion involves randomly selecting a syllable from a sentence, choosing a character from the Tibetan alphabet, and inserting it at a random position within the selected syllable.
The pseudo-code is presented in  Algorithm~\ref{alg:char_random_insert}.
\begin{algorithm}[ht]
    \caption{char\_random\_insert}\label{alg:char_random_insert}
    \begin{algorithmic}[1]
    \STATE \textbf{Input:} List of syllables \texttt{syllables}, List of Tibetan alphabet \texttt{alphabet}
    \STATE \texttt{selected\_char} = Random(\texttt{alphabet})
    \STATE Randomly select insert index: \texttt{i} = Random(0, len(syllables)-1)
    \STATE Insert \texttt{element} at index \texttt{i} in \texttt{syllables}: \texttt{syllables.insert(i, selected\_char)}
    \STATE \textbf{Return} \texttt{syllables}
    \end{algorithmic}
\end{algorithm}

Character-level case substitution involves randomly selecting a character that is a root letter with both uppercase and lowercase forms and replacing it with its alternate case.
The pseudo-code is presented in  Algorithm~\ref{alg:char_case_substitution}.
\begin{algorithm}[ht]
    \caption{char\_case\_substitution}\label{alg:char_case_substitution}
    \begin{algorithmic}[1]
    \STATE \textbf{Input:} List of syllables \texttt{syllables}
    \STATE Randomly select a syllable: \texttt{syllable} = Random(\texttt{syllables})
    \STATE Randomly Select a character in \texttt{syllable}: \texttt{char} = Random(\texttt{syllable})
    \IF{\texttt{char} in LOWER}
        \STATE Replace \texttt{char} with \texttt{lower2upper(char)}
    \ELSIF{\texttt{char} in UPPER}
        \STATE Replace \texttt{char} with \texttt{upper2lower(char)}
    \ELSE
        \STATE Keep \texttt{char}
    \ENDIF
    \STATE Update \texttt{syllables} at the respective syllable indices
    \STATE \textbf{Return} \texttt{syllables}
    \end{algorithmic}
\end{algorithm}

Character-level homoglyph substitution involves randomly selecting a character with multiple homoglyphs and replacing it with one of its variants.
The pseudo-code is presented in  Algorithm~\ref{alg:char_homomorphic_substitution}.
\begin{algorithm}[ht]        \caption{char\_homomorphic\_substitution}\label{alg:char_homomorphic_substitution}
    \begin{algorithmic}[1]
    \STATE \textbf{Input:} List of syllables \texttt{syllables}
    \STATE Find syllables with homoglyph characters: \texttt{syllables\_sel}
    \IF{\texttt{syllables\_sel} is empty}
        \STATE \textbf{Return} \texttt{syllables}
    \ENDIF
    \STATE Randomly select a element from \texttt{syllables\_sel}: \texttt{syllable}
    \STATE Find the index of selected syllable: \texttt{i}
    \STATE Randomly select a indice of replaceable character in \texttt{syllable}: \texttt{j}
    \STATE Select character: \texttt{char} = \texttt{syllable[j]}
    \STATE Replace \texttt{char} with a random homoglyph from \texttt{HOMOMORPHIC\_LETTER}
    \STATE Update \texttt{syllables[i]}
    \STATE \textbf{Return} \texttt{syllables}
    \end{algorithmic}
\end{algorithm}

Adjacent-syllabic character transposition refers to the process of randomly selecting a syllable within a sentence and swapping the positions of two characters inside that syllable.
The pseudo-code is presented in  Algorithm~\ref{alg:char_inner_syllable_transposition}.
\begin{algorithm}[ht]
    \caption{adjacent\_syllable\_char\_transposition}\label{alg:char_inner_syllable_transposition}
    \begin{algorithmic}[1]
    \STATE \textbf{Input:} List of syllables \texttt{syllables}
    \STATE Find long syllables (length > 2): \texttt{long\_syllables}
    \IF{\texttt{long\_syllables} is empty}
        \STATE \textbf{Return} \texttt{syllables}
    \ENDIF
    \STATE Randomly select a long syllable from \texttt{long\_syllables}
    \STATE Find the index of selected syllable: \texttt{syllable\_index}
    \STATE Randomly select two indices \texttt{idx1} and \texttt{idx2}
    \STATE Swap characters at \texttt{idx1} and \texttt{idx2}
    \STATE Update \texttt{syllables[i]}
    \STATE \textbf{Return} \texttt{syllables}
    \end{algorithmic}
\end{algorithm}

Inter-syllabic character transposition refers to the process of randomly selecting two adjacent syllables in a sentence and exchanging one character between them.
The pseudo-code is presented in  Algorithm~\ref{alg:char_near_syllable_transposition}.
\begin{algorithm}[ht]
    \caption{inter\_syllable\_char\_transposition}\label{alg:char_near_syllable_transposition}
    \begin{algorithmic}[1]
    \STATE \textbf{Input:} List of syllables \texttt{syllables}
    \IF{length of \texttt{syllables} < 2}
        \STATE \textbf{Return} \texttt{syllables}
    \ENDIF
    \STATE Randomly select two adjacent syllables: \texttt{syll1} and \texttt{syll2}
    \STATE Randomly select a character from \texttt{syll1} and \texttt{syllable2}
    \STATE Swap characters between \texttt{syll1} and \texttt{syll2}
    \STATE Update \texttt{syllables} at the respective syllable indices
    \STATE \textbf{Return} \texttt{syllables}
    \end{algorithmic}
\end{algorithm}

Syllable-level random deletion involves randomly removing a syllable from a sentence. In our semi-masked approach, a corresponding label is generated that retains the uncorrupted parts of the text while replacing the deleted syllables with a [MASK] token.
The pseudo-code is presented in  Algorithm~\ref{alg:syllable_random_delete}.
\begin{algorithm}[ht]
    \caption{syllable\_random\_delete}\label{alg:syllable_random_delete}
    \begin{algorithmic}[1]
    \STATE \textbf{Input:} List of syllables \texttt{syllables}
    \STATE Copy \texttt{syllables} into \texttt{syllables\_mask}
    \STATE Randomly select \texttt{idx} from range \texttt{[0, len(syllables) - 1]}
    \STATE Remove element at \texttt{idx} from \texttt{syllables}
    \STATE Set \texttt{syllables\_mask[idx]} to \texttt{[MASK]}
    \STATE \textbf{Return} \texttt{syllables, syllables\_mask}
    \end{algorithmic}
\end{algorithm}

Syllable-level random transposition refers to the process of randomly selecting two syllables within a sentence and swapping their positions.
The pseudo-code is presented in  Algorithm~\ref{alg:syllable_random_substitution}.
\begin{algorithm}[ht]
    \caption{syllable\_random\_transposition}\label{alg:syllable_random_substitution}
    \begin{algorithmic}[1]
    \STATE \textbf{Input:} List of syllables \texttt{syllables}
    \IF{\texttt{len(syllables)} $<$ 2}
        \STATE \textbf{Return} \texttt{syllables}
    \ENDIF
    \STATE Randomly select \texttt{i, j} from the range \texttt{[0, len(syllables))}, ensuring they are different
    \STATE Swap \texttt{syllables[i]} and \texttt{syllables[j]}
    \STATE \textbf{Return} \texttt{syllables}
    \end{algorithmic}
\end{algorithm}

Syllable-level random merging refers to the process of randomly selecting two adjacent syllables in a sentence and merging them into a single syllable.
The pseudo-code is presented in  Algorithm~\ref{alg:syllable_random_merging}.
\begin{algorithm}[ht]
    \caption{syllable\_random\_merging}\label{alg:syllable_random_merging}
    \begin{algorithmic}[1]
    \STATE \textbf{Input:} List of syllables \texttt{syllables}
    \IF{\texttt{len(syllables)} $<$ 2}
        \STATE \textbf{Return} \texttt{syllable\_list}
    \ENDIF
    \STATE Randomly select \texttt{idx} from range \texttt{[0, len(syllables) - 2]}
    \STATE Merge \texttt{syllable\_list\_copy[idx]} and \texttt{syllable\_list\_copy[idx + 1]}
    \STATE Remove \texttt{syllables[idx + 1]}
    \STATE \textbf{Return} \texttt{syllables}
    \end{algorithmic}
\end{algorithm}

\end{document}